\documentclass{article}

\usepackage{arxiv}

\usepackage{times}
\usepackage[numbers,sort,sectionbib]{natbib}
\usepackage{url}
\usepackage{latexsym}
\usepackage[numbers,sort,sectionbib]{natbib}
\usepackage{graphicx}

\makeatletter
\renewcommand\@biblabel[1]{#1.}
\usepackage{wrapfig}
\usepackage{algorithm}
\usepackage{algpseudocode}
\usepackage{tabularx}
\usepackage{amssymb}
\usepackage{hyperref}

\title{A Review of the Challenges with Massive Web-mined Corpora Used in Large Language Models Pre-Training}

\author{ \href{https://orcid.org/0000-0001-8646-3345}{\includegraphics[scale=0.06]{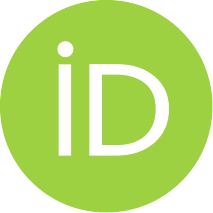}\hspace{1mm}Michał~Perełkiewicz} \\
	National Information Processing Institute \\
    al. Niepodległości 188B, Warsaw, Poland \\
	\texttt{mperelkiewicz@opi.org.pl} \\
	\And
	\href{https://orcid.org/0000-0002-6108-2711}{\includegraphics[scale=0.06]
    {orcid.pdf}\hspace{1mm}Rafał~Poświata} \\
	National Information Processing Institute \\
    al. Niepodległości 188B, Warsaw, Poland \\
	\texttt{rposwiata@opi.org.pl} \\
}

\renewcommand{\shorttitle}{\textit{arXiv} Template}

\hypersetup{
pdftitle={A template for the arxiv style},
pdfsubject={q-bio.NC, q-bio.QM},
pdfauthor={David S.~Hippocampus, Elias D.~Striatum},
pdfkeywords={First keyword, Second keyword, More},
}

\begin{document}
\maketitle

\begin{abstract}
This article presents a comprehensive review of the challenges associated with using massive web-mined corpora for the pre-training of large language models (LLMs). This review identifies key challenges in this domain, including challenges such as noise (irrelevant or misleading information), duplication of content, the presence of low-quality or incorrect information, biases, and the inclusion of sensitive or personal information in web-mined corpora. Addressing these issues is crucial for the development of accurate, reliable, and ethically responsible language models. Through an examination of current methodologies for data cleaning, pre-processing, bias detection and mitigation, we highlight the gaps in existing approaches and suggest directions for future research. Our discussion aims to catalyze advancements in developing more sophisticated and ethically responsible LLMs.
\end{abstract}


\section{Introduction}

The advent of large language models (LLMs) has heralded a new era in natural language processing (NLP), offering capabilities that range from sophisticated text generation to nuanced language understanding. These advancements have been propelled by significant improvements in model architectures, algorithms, and, crucially, the availability of extensive datasets for training. Given the data-intensive nature of these models, the quest for high-quality, diverse, and substantial datasets has become paramount. In this context, massive web-mined corpora have emerged as a vital resource, offering an abundance of textual data that mirrors the vastness and variety of human language and interaction  \cite{Jakubicek2020, 10.5555/3455716.3455856, penedo2023refinedweb, touvron2023llama}.

The internet, with its exponential growth and dynamic content, presents a near-infinite source of text data, spanning every conceivable topic, language, and style. This richness makes web-mined data an attractive foundation for training LLMs, aiming to equip them with a broad understanding of language and its applications. However, the use of such data is not without its challenges. The process of web mining—extracting data from websites—entails navigating a complex landscape of technical, legal, ethical, and quality-related issues \citep{Veres2021LargeLM, wenzek2020ccnet, elazar2024whats, Dodge2021DocumentingLW, gao2020pile}.

By critically examining the use of web-mined corpora in the pre-training of LLMs, this article contributes to a nuanced understanding of the current landscape and future directions in large-scale language model development.

\section{The Nature of Web-mined Corpora}

The internet serves as an abundant source of data with great potential for use in training advanced artificial intelligence models. Data collections such as Common Crawl\footnote{https://commoncrawl.org/}, consisting of textual data scraped from the web, are a treasure trove of linguistic diversity and richness. This section briefly describes the characteristics of such corpora.

\subsection{Defining Web-mined Corpora}

At its core, a web-mined corpus is a collection of textual data that has been extracted from the internet. This includes a wide range of content such as websites, blogs, forums, social media posts, and other digital texts. The process of web mining involves the automated scraping of this content, followed by stages of cleaning and pre-processing to prepare the data for use in machine learning applications. The primary allure of web-mined corpora lies in their ability to reflect the multifaceted nature of human language and communication as manifested online.

\subsection{Scale and Diversity}

Web-mined corpora can span billions of words, encompassing numerous languages, dialects, genres, and domains. This scale and variety are unparalleled compared to traditionally curated datasets, providing LLMs with a broad spectrum of linguistic patterns, contexts, and nuances to learn from. Such extensive training data is crucial in developing models that can understand and generate human-like text across a wide array of applications \citep{penedo2023refinedweb, together2023redpajama}.

However, the sheer volume of data available through web mining also introduces significant challenges. The task of filtering, cleaning, and organizing this data into a coherent corpus that is suitable for training is non-trivial. Issues such as duplicate content, irrelevant or low-quality text, and the presence of sensitive or personal information necessitate meticulous pre-processing and ethical considerations \citep{wenzek2020ccnet, kreutzer-etal-2022-quality, Kaddour2023ChallengesAA, Hadi_2023}.

\subsection{Widely used Web-mined Corpora}

We describe several pivotal web-mined corpora widely employed in the training of LLMs, highlighting their respective contributions to the field:

\textbf{Common Crawl} is an open repository of web crawl data collected over years. The Common Crawl repository provides around 20TB of scraped text data each month. Due to the wide coverage of diverse domains and languages, as well as the size of the collected data, Common Crawl's archives have become a key resource in the corpus of pre-training major language models in recent years \citep{touvron2023llama, soldaini2024dolma, 10.5555/3455716.3455856}.

\textbf{C4} (Colossal Clean Crawled Corpus) is a collection of about 750GB of English-language text sourced from the April 2019 snapshot of Common Crawl. The cleaning process involves removing duplicate content, boilerplate text like headers and footers, and filtering out low-quality content. The C4 corpus was originally created to train the T5 model \citep{10.5555/3455716.3455856}. A multilingual variant of the C4 corpus -- \textbf{mC4} contains documents collected from Common Crawl through August 2022 and was used to train mT5 \citep{xue-etal-2021-mt5} and umT5 models \citep{chung2023unimax}.

\textbf{The RefinedWeb} is another massive corpus derived from the highly curated and deduplicated Common Crawl data. It was used to train the Falcon LLM model \citep{penedo2023refinedweb}. Publicly released version of this corpus contain extracted 600B tokens. 

\textbf{OSCAR} is a multilingual corpus based on Common Crawl \citep{OrtizSuarezSagotRomary2019}. It contains a length filter for improving data quality that filters out documents with short sentences. They also
annotate the data with different labels, such as the language of the document, adult content, and language identification, which they use for different analyses. The dataset used for training multilingual models such as BART \citep{lewis-etal-2020-bart} incorporates 138 GB of text.

\textbf{WebText}: Developed by OpenAI, the corpus consists of a filtered selection of web page texts aimed at excluding low-quality content. The curation process behind WebText is designed to optimize the informational value of the data, supporting the development of LLMs with refined understanding and generative capabilities \citep{Radford2019LanguageMA}. \textbf{OpenWebText}  is an open-source reproduction \citep{Gokaslan2019OpenWeb} of the corpus used to train the GPT-2 model \citep{Radford2019LanguageMA}. The corpus is essentially a collection of text extracted from web pages, specifically gathered from URLs shared on Reddit and filtered based on the number of upvotes. This method aimed to replicate the dataset used by OpenAI for training its language models, particularly focusing on high-quality and popular content as determined by Reddit users.

\textbf{The Pile} is a comprehensive and diverse dataset created for training large language models in the field of natural language processing (NLP). One of the key components of The Pile is Pile-CC, which stands for the Common Crawl portion of The Pile \citep{gao2020pile}. Models such as GPT-J \citep{gpt-j}, GPT-neo \citep{black-etal-2022-gpt} and Pythia \citep{biderman2023pythia} were trained on this dataset.

\textbf{RedPajama-Data-v2} is an open dataset for training large language models. The dataset includes over 100B text documents coming from 84 Common Crawl snapshots and processed using the CCNet pipeline \citep{wenzek2020ccnet}. the corpus was used to train the LLaMa model \citep{touvron2023llama}.

\section{Challenges in Using Web-mined Corpora for LLM Pre-Training}

In this chapter, we present an overview of the inherent challenges and considerations involved in utilizing web-mined corpora for the pre-training of LLMs. We explore the complexities of data quality, representativeness, ethical concerns and reliability of model evaluations, shedding light on the obstacles and potential solutions in harnessing the vast resources of the internet for advancing AI research.

\subsection{Data Quality and Noise}

Due to their raw nature, web-mined corpora are characterized by low-quality text data - insufficient for direct use in training LLMs \citep{gao2020pile, NEURIPS2020_1457c0d6}. Available in the literature, analyses of corpora derived from the Common Crawl indicate a positive impact of LLMs training on largely filtered data considered to be of low quality \citep{wenzek2020ccnet, marion2023more, penedo2023refinedweb, NEURIPS2020_1457c0d6}. The text quality assessment and data cleaning methods used in the works analyzed include: 1) rules and heuristic methods for filtering poor quality texts, 2) pre-trained text quality classifiers, 3) use of statistical models for automatic language identification, 4) filtering of poor quality content based on n-grams, lightweight language models.

\textbf{Rules and Heuristic Methods for Filtering Poor Quality Texts.} These methods for filtering low-quality texts described in the literature focus on cleaning the data based on a set of established, explainable rules for assessing text quality. They usually involve filtering content with poor punctuation accuracy, documents that are too short or long (in terms of the number of characters), filtering texts containing HTML tags or JavaScript scripts and "lorem ipsum" phrases, removing privacy policy content such as: "terms of use", "privacy policy", "cookie policy", "uses cookies", "use of cookies", "use cookies", removing content containing words from the lists of prohibited words; as well as rules such as removing documents in which more than 90\% of lines start with bullets, more than 30\% of sentences end with ellipses, etc. \citep{together2023redpajama, wenzek2020ccnet, penedo2023refinedweb, 10.5555/3455716.3455856, hernandez2022scaling}. Methods are also used to define sets of text quality features (called quality signals), such as: text length, share of capital letters, share of non-alphanumeric characters, share of unique words, share of source code in the text, etc., followed by filtering low-quality text based on these defined characteristics \citep{gao2020pile, 10.5555/3454287.3454804, 10.5555/3455716.3455856, wenzek2020ccnet}.

\textbf{Pre-Trained Text Quality Classifiers.} The use of pre-trained text quality classification/regression models is a common element of low-quality content filtration methods described in the literature \citep{NEURIPS2020_1457c0d6, laippala-etal-2020-web, penedo2023refinedweb, wenzek2020ccnet}. For training classifier models, corpora that are considered to be of good quality are usually used, such as Wikipedia text corpora, BookCorpus\footnote{https://github.com/soskek/bookcorpus} and other curated dataset \citep{NEURIPS2020_1457c0d6}.

\textbf{Statistical Models for Automatic Language Identification.} Using statistical, \newline lightweight text language identification models such as langid.py\footnote{https://github.com/saffsd/langid.py} \citep{grave-etal-2018-learning}, fasttext\footnote{https://fasttext.cc/docs/en/language-identification.html} \citep{gao2020pile}, langdetect\footnote{https://pypi.org/project/langdetect/} \citep{10.5555/3455716.3455856} is another method of filtering qualitative data from CommonCrawl corpora \citep{penedo2023refinedweb, OrtizSuarezSagotRomary2019} described in the literature.

\textbf{Probabilistic N-grams Language Models.} Filtering low-quality content based on n-gram lightweight language models trained on good-quality data collections is an element of data cleaning and filtration processes described in the literature \citep{marion2023more}. Pipelines like CCNet \citep{wenzek2020ccnet} employ statistical KenLM models to assess text perplexity during the data filtering phase.

The poor quality of data collected in web corpora is also largely due to the difficulty of extracting the main content from websites. Contamination caused by text extraction from unwanted HTML elements such as menus, headers, footers, breadcrumbs, advertisements, copyright notices, spam largely affects the performance of language models \citep{barbaresi-2016-efficient, Kristoffersen2017CommonCW, book-Paquot-2020, laippala-etal-2020-web, kaddour2023minipile}. Analysis of the most frequently occurring \textit{n}-grams conducted on OpenWebText, C4, mC4, OSCAR, The Pile and RedPajama indicates a large share of sequences of repeated punctuation characters such as dashes, question marks, dots; HTML tags, and content derived from the extraction of such page elements like headers, menus etc \cite{elazar2024whats}.

Comparative analyses of the quality of models trained on raw and filtered pre-training corpora confirm the negative impact of low-quality texts in web corpus and point to the need for methods of filtering high-quality data from web corpora in the process of LLMs pre-training \cite{penedo2023refinedweb, kaddour2023minipile}.

\subsection{Bias and Representativeness}
\label{bias}

A problem often raised in the literature is the overrepresentation of English-language texts and the uneven distribution of data in terms of analyzed domains and website addresses, as well as the geolocation of IP addresses of URLs and the characteristics of users creating web content \citep{Dodge2021DocumentingLW}.

\textbf{Language.} Statistics of data collected in Common Crawl's CC-MAIN-2023-50 archive show that the share of content identified as English-language texts was about 44.42\% of the content. The second most popular language, German, accounted for 5.44\% in the examined archive, while the Polish language was identified in 1.75\% of the data. Only 14 of the 160 languages surveyed had a share greater than 1\%. This overrepresentation of English in Common Crawl data allows for the construction of huge derivative corpora for English, but can be a problem when building databases for less represented languages.

\textbf{TLDs and URLs.} The analysis of top-level domains (TLDs) in the C4 corpus \citep{10.5555/3454287.3454804} showed an overrepresentation of data from the \textit{.mil} domain intended for government entities and organizations that are part of the The United States Department of Defense \citep{Dodge2021DocumentingLW}. Patent search engine URLs such as \texttt{patents.google.com} and \texttt{patents.com} \citep{Dodge2021DocumentingLW} are also overrepresented in the C4 corpus (respectively most scraped domain and 10th most scraped). The RedPajama corpus contains approximately 12\% of data comes from the \texttt{arxic.ora} domain and approximately 7\% data from the \texttt{stackoverflow.com} domain \citep{ elazar2024whats}. This kind of bias does not evenly reflect the diversity of content found on the web.

\textbf{Geolocation.} The Common Crawl data is naturally overrepresented by the Internet user population -- younger, English-speaking people from developed regions of the world \citep{luccioni-viviano-2021-whats}.
An analysis of the C4 corpus in terms of the host location of the analyzed sites indicates that 51.3\% of them are hosted in the United States. The countries with the estimated second, third and fourth largest English-speaking populations - India, Pakistan, Nigeria and the Philippines\footnote{\url{https://en.wikipedia.org/wiki/List\_of\_countries\_by\_English-speaking\_population}} - have only 3.4\%, 0.06\%, 0.03\%, 0.1\% share of URLs in the studied corpus, despite having a large population, including a large share of English speakers \citep{Dodge2021DocumentingLW}.

\textbf{Gender and Age of Web Content Creators.} The study \citep{10.1145/3442188.3445922} highlights the drawbacks of the WebText and OpenWebText corpus building method. It assumes scraping content from links from the Reddit platform\footnote{https://www.reddit .com/}, where according to a 2016 Pew Internet Research report, 67\% of users are male, and 64\% of users are between the ages of 18 and 29\footnote{\url{https://www.pewresearch.org/internet/2016/11/11/social-media-update-2016/}}. Inequality in the distribution of users by gender has also been pointed out on Wikipedia, where only 8.8-15\% of editing users are women.

\subsection{Ethical Considerations} 

Undesirable content in web corpora includes hate speech, sexually explicit content, racist and xenophobic content, sensitive data such as personal data, contact details of private persons, etc. The presence of this type of content in pre-training corpora affects the risk of the model generating offensive content as well as the trained model duplicating personal and personally identifiable information about specific individuals \citep{elazar2024whats, Carlini2022QuantifyingMA, chen2023language}.

The research \citep{luccioni-viviano-2021-whats} conducted on 115 GB of randomly selected English-language data from Common Crawl (November/December 2020 snapshot) showed the presence of hate speech at levels ranging from 4.02\% to 6.38\%. The most frequently cited examples of hate speech were racial slurs, racial supremacy and racially motivated conspiracy theories. In the same study, the presence of sexually explicit content was estimated at approximately 2.36\% in the examined data.

Another research \citep{gehman-etal-2020-realtoxicityprompts} on the occurrence of undesirable content\footnote{Understood as offensive, xenophobic and racist content.} in the WebText and OpenWebText corpora found that both corpora contain toxic content at a level of 4.3\% and 2.1\%, respectively. These rates are consistent with the \citep{founta2018large} study, which found that the prevalence of offensive or toxic content roughly ranges between 0.1\% and 3\%, and suggest that these corpora merely reflect "natural'' toxicity indicators.

The analysis of the C4 and The Pile corpora \citep{subramani-etal-2023-detecting} showed the presence of Personally Identifiable Information (PII) data such as telephone numbers, e-mail addresses, bank account numbers, social security numbers (SSN), IP addresses, IBAN numbers.

The authors of the analysis of the C4 \citep{Dodge2021DocumentingLW} corpus drew attention to too aggressive filtering of undesirable content used, resulting in the removal of too much content regarding sexual orientation, content related to science, medicine and health, as well as content written in African American English and Hispanic-aligned English languages.


\subsection{Duplication of Content}

Content duplication in pre-training corpora negatively affects the quality of trained models by 1) increasing the risk of remembering repetitive token sequences, 2) increasing the risk of documents from the training corpus appearing in the validation and testing sets, 3) increasing the training time and cost, 4 ) in some cases increases model perplexity \citep{hernandez2022scaling, kaddour2023minipile, lee-etal-2022-deduplicating, Kaddour2023ChallengesAA, soldaini2024dolma}.

The analysis of duplicates in the Common Crawl collections carried out as part of the RedPajama-V2 \citep{together2023redpajama} project showed that approximately 40\% of the data in 84 archives were exact duplicates. The C4 \citep{10.5555/3455716.3455856} and RealNews \citep{zellers2020defending} corpora according to \citep{lee-etal-2022-deduplicating} research contain close duplicates\footnote{Close duplicates in the meaning of the measure Jaccard with a similarity threshold of 0.8. The research used the MinHash algorithm.} at the level of 3.04\% and 13.63\%, respectively, despite the use of deduplication in the corpus creation process. For the validation sets of these two corpora, 4.60\% and 14.35\% of the texts were found to be close duplicates with the texts from the training sets, respectively.

The CCNet project's deduplication of text paragraphs (pre-processed to include reducing all characters to lowercase, replacing numeric values with "0" and removing punctuation and diacritics) extracted from the Common Crawl archive showed a paragraph duplication level of 70\% in the analyzed data \citep{wenzek2020ccnet}. Examination of exact duplicates at the line level of documents from the CommonCrawl and Wikipedia archives (data extracted from Wikipedia's public archives\footnote{https://dumps.wikimedia.org/}) indicated a duplicate occurrence of 37\% and 21\% respectively\citep {grave-etal-2018-learning}. Previous research conducted on Common Crawl archives from 2012-2013 showed duplicate lines in documents at the level of 80\% \citep{buck-etal-2014-n}. Other web corpora with a high proportion of duplicates include The Pile, LAION-2B-en, RedPajama, OSCAR according to the analysis \citep{elazar2024whats}.

\subsection{Low-Resources Languages}

In addition to the low availability of texts for low-resource languages (see Chapter \ref{bias}) in web corpora, a problem reported in the literature for this category of languages is low quality due to the high proportion of automatically translated texts in the scraped data \citep{kreutzer-etal-2022-quality}.

\textbf{Automatically Translated Content. } A large amount of non-English texts in the Common Crawl corpus is generated by generative machine learning models, especially machine translation models \citep{arasezhou2013, Rarrick2011MTDI, Mehmood2017UnderstandingRC, thompson2024}. In analyzes of non-English texts in Common Crawl data, errors typical of machine translation models are pointed out, such as incorrect translation of dates as well as names, surnames or location names, resulting in the occurrence of incorrect information that can be interpreted by the model as facts \citep{kreutzer-etal-2022-quality}. Other errors resulting from incorrect translations most often concern changes in numerical values (stock exchange information, exchange rates, percentages, prices, proper names such as company names, city names, etc.) \citep{Dodge2021DocumentingLW, kreutzer-etal-2022-quality}.

\subsection{Benchmark Data Contamination}

Benchmark data contamination occurs when the data used to train an LLM overlaps with the datasets employed for testing or evaluating the model's performance. Given the vast expanse of web-mined corpora, unintentional inclusion of benchmark data in the training set is not only possible but likely. This contamination can artificially inflate the model's performance metrics, leading to an overestimation of its true capabilities. As such, it is important to track contamination \citep{jacovi-etal-2023-stop}.

The research analyzing The Pile, C4, RedPajama, and OSCAR corpora found that RedPajama is the most contaminated dataset among the studied ones (considering PromptSource benchmarks). For eight out of the 15 corpora from PromptSource, its contamination rate is above 50\% with fully contaminated in the case of COPA \citep{elazar2024whats}. The Pile’s contamination rates are lower, but it is also contaminated with a few datasets, such as WSC and WIC, which were included in the SuperGLUE evaluation benchmark \citep{NEURIPS2019_4496bf24}. The study \citep{Dodge2021DocumentingLW} indicates significant contamination of the C4 web corpus with benchmark datasets LAMA T-REx derived from and Google RE derived from the Google-RE corpus\footnote{https://code.google.com/archive/p/relation-extraction-corpus/}.






\section{Conclusion}

The use of massive web-mined corpora has become an essential step in training large language models. While the web offers a huge amount of data, it also introduces complexities related to data quality, ethical considerations, data biases and benchmark data contamination. Addressing these challenges requires a multifaceted approach, incorporating robust data cleaning techniques, ethical data use practices, and efforts towards creating comprehensive frameworks for bias mitigation as well as development of new benchmark datasets and evaluation metrics. As we navigate these issues, the evolution of web content and technology continues to shape the path forward.

\bibliographystyle{./splncs04}
\bibliography{refs}






\end{document}